%% file: root.tex
\documentclass[letterpaper, 10 pt, conference]{ieeeconf}
\IEEEoverridecommandlockouts                              

\usepackage{graphicx}
\usepackage{comment}
\usepackage{amsmath,amssymb} 
\usepackage{color}
\usepackage{epsfig}
\usepackage{graphicx}
\usepackage{amsmath}
\usepackage{amssymb}
\usepackage{pifont}

\usepackage{color}
\usepackage{xcolor}
\usepackage{epstopdf,cite,subfigure}
\usepackage{algorithm,algpseudocode}
\usepackage{pdfpages}
\usepackage{multirow}
\usepackage{rotating}
\usepackage{array}
\usepackage{epsf}
\usepackage{graphics}
\usepackage{makecell}
\usepackage{adjustbox}  
\usepackage{xpunctuate}
\usepackage{foreign}
\usepackage{pbox}
\usepackage{tabularx,booktabs,calc}
\usepackage{hhline}
\usepackage{stackengine}
\usepackage{mathtools}
\DeclarePairedDelimiter\norm{\lVert}{\rVert}
\usepackage{amsmath}
\usepackage{romannum}

\title{\LARGE \bf
Joint Representation of Temporal Image Sequences and Object Motion  for Video Object Detection
}

\author{Junho Koh$^{1}$*, Jaekyum Kim$^{1}$*, Younji Shin$^{1}$, Byeongwon Lee$^{2}$, Seungji Yang$^{2}$ and Jun Won Choi$^{1}$
\thanks{* indicates equal contribution.}
\thanks{$^{1}$	
Department of Electrical Engineering, Hanyang University, Seoul 04763, Korea}
\thanks{$^{2}$ T3K Vision AI Product Center of Excellence, SK Telecom, Seongnam 13487, Korea}
\thanks{{\tt\small\{jhkoh,jkkim,yjshin\}@spa.hanyang.ac.kr}}
\thanks{{\tt\small \{bwon.lee,yangs\}@sk.com}}
\thanks{{\tt\small junwchoi@hanyang.ac.kr}}

}

\begin{document}

\maketitle
\thispagestyle{empty}
\pagestyle{empty}

\begin{abstract}
     In this paper, we propose a new video object detector (VoD) method referred to as {\it temporal feature aggregation and motion-aware VoD} (TM-VoD), which produces a joint representation of temporal image sequences and object motion. The proposed TM-VoD aggregates visual feature maps extracted by convolutional neural networks applying the temporal attention gating and spatial feature alignment. This temporal feature aggregation is performed in two stages in a hierarchical fashion. In the first stage, the visual feature maps are fused at a pixel level via {\it gated attention model}. In the second stage, the proposed method aggregates the features after aligning the object features using {\it temporal box offset calibration} and weights them according to the cosine similarity measure. The proposed TM-VoD also finds the representation of the motion of objects in two successive steps. The pixel-level motion features are first computed based on the incremental changes between the adjacent visual feature maps. Then, box-level motion features  are obtained from both the region of interest (RoI)-aligned pixel-level motion features and the sequential changes of the box coordinates. Finally, all these features are concatenated to produce a joint representation of the objects for VoD. The experiments conducted on the {\it ImageNet VID dataset} demonstrate that the proposed method outperforms existing VoD methods and achieves a performance comparable to that of state-of-the-art VoDs. 
\end{abstract}



\section{INTRODUCTION}

  In the field of robot vision,  the performance of object detectors, including SSD \cite{SSD}, YOLO \cite{YOLO}, RetinaNet \cite{RetinaNet}, Faster R-CNN  \cite{Faster_RCNN}, and Mask R-CNN \cite{Mask_RCNN},   has been improved dramatically since convolutional neural network (CNN) \cite{alexnet, vgg, resnet, mobilenet} have been adopted for feature extraction of images.
      These well-known object detectors detect the objects based on a single image. When object detection is performed on video data that contains a sequence of image frames, the traditional approach is to perform detection for each image frame  and to associate objects across frames in the subsequent object tracking stage. However, this approach does not exploit the temporal information in the image sequence, thereby limiting the detection performance.  
     In addition, video images often suffer from degraded image quality due to motion blur, camera defocusing, anomalous poses, and object occlusion. Since this gives inconsistent detection results over time, and consequently burdens the object trackers, the object detectors should be designed to exploit temporal information to achieve the robust performance.
    
    Recently, object detectors, referred to as {\it video object detectors} (VoD), have been proposed, which use multiple consecutive video frames for object detection. 
     Thus far, various VoD methods have been proposed in the literature \cite{FGFA,stsn,stmn,psla,selsa,rdn,mega}. 
     In \cite{FGFA, stmn, stsn, psla}, CNN feature maps were fused to produce an enhanced representation of objects for object detection. In particular, the methods in \cite{rdn,selsa,mega} associated the object proposals found in each video frame and fused the associated features to enhance the quality of the object features. In \cite{detecttotrack} and \cite{manet}, the motion of objects and the variation of camera position and angle were exploited to extract the representation of the moving objects. 
      

 In this paper, we present a novel VoD algorithm, referred to as {\it temporal feature aggregation and motion-aware VoD (TM-VoD)},  which can construct  robust and reliable features on objects using image sequences of finite length.
 We aim to design a VoD algorithm that achieves the following two objectives of VoD.
  First, the VoD algorithm should aggregate common, yet diverse representations of objects over multiple video frames. Since the location and the quality of object features change in time, the aggregation strategy should be adapted to such temporal variations.
Next, the VoD algorithm should exploit the temporal motion patterns of objects to find rich and discriminative representations. Since objects of different classes exhibit distinctive motion patterns, the respective motions provide useful contextual cues for identifying the objects better. 
 
\begin{figure*}[t]
    	\centering
        \centerline{\includegraphics[width=0.88\textwidth]{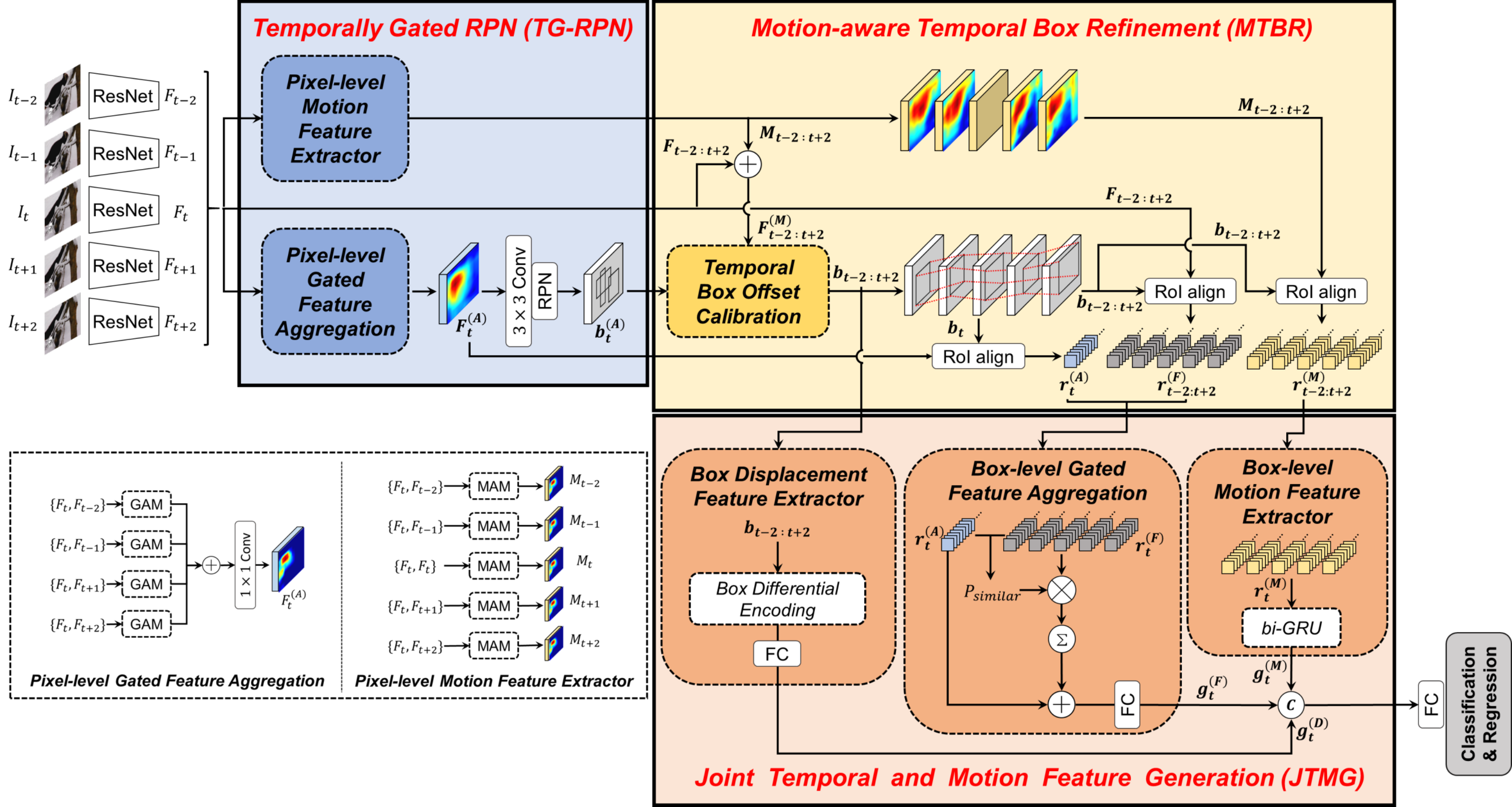}}
    	\caption {\textbf{Overall architecture of  TM-VoD}: The TM-VoD consists of three main blocks. TG-RPN generates the region proposals for the current frame based on the selectively aggregated feature map and the pixel-level motion features. 
      MTBR predicts the coordinates of the box proposals linked over all video frames via TBOC. Both visual and motion features are pooled based on the predicted boxes from TBOC via the {\it RoI alignment} block. Finally, JTMG produces the final box-level object features by generating the selectively fused visual features and the box-level motion features.  
                    }
    	\label{overall}
    \end{figure*}

  The proposed TM-VoD method detects objects based on $M$ past images, $N$ future images, and present image as illustrated with the setup $N=M=2$ in Fig. \ref{overall}.  First, the TM-VoD fuses the visual feature maps obtained by the CNN backbone networks. To maximize the effect of feature aggregation, TM-VoD aligns and weights the feature maps to be aggregated in two stages. In the first stage, the {\it pixel-level gated feature aggregation} performs a weighted aggregation of the CNN feature maps based on their relevance to the detection task at hand. In the second stage, the box proposals obtained by the region proposal network (RPN) are aligned by {\it temporal box offset calibration} (TBOC) and weighted according to the cosine similarity between the present and adjacent frame features.
  TM-VoD also finds the representation of object motion in two stages. In the first stage, the pixel-level motion features are obtained by capturing incremental changes of adjacent visual feature maps. In the second stage, box-level motion features are extracted from the region of interest (RoI)-aligned motion features and the sequence of the corresponding box coordinates. Finally, both temporally aggregated features and box-level motion features are merged to generate a joint representation of the objects.  Note that the entire network is end-to-end trainable.

    The key contributions of our study are summarized as follows.
     \begin{itemize}

         \item We propose a new VoD method, which exploits both the temporal redundancy of object features over adjacent video frames and the contextual information captured in the motion of objects. We particularly focus on aggregating only the relevant and well-aligned regions of the visual feature maps, extracting the effective representation of object motion from the image sequence, and integrating them to the VoD.
     Note that the feature aggregation and motion feature extraction are performed sequentially in a hierarchical manner, both at the pixel level and at the box level.
       

         \item We propose an efficient box regression method for aligning box proposals over multiple video frames. 
          Instead of associating all pairs of box proposals over adjacent frames \cite{selsa, rdn, mega}, 
          the proposed method uses the initial box proposals obtained by the RPN as anchors and predicts the box offsets relative to the anchors for all video frames.
       
         \item  We evaluate the performance of the proposed method on the publicly available video object detection dataset, {\it ImageNet VID dataset} \cite{imagenet}. Our experiments demonstrate that the proposed ideas offer significant performance improvement over the baseline algorithms. Furthermore, the proposed TM-VoD method  outperforms existing VoD methods and achieves a performance comparable to that of state-of-the-art methods. 
         \item The source code will be publicly available. 
    \end{itemize}


\section{RELATED WORK}
    
    Object detection techniques based on still images \cite{RCNN, Fast_RCNN, Faster_RCNN, RFCN, Mask_RCNN, YOLO, SSD, RetinaNet} have been rapidly advanced owing to the use of CNN. However, the performance of these object detectors is limited because temporal information in image sequences is not exploited. VoD methods have been recently proposed, which uses the sequence of the video frames for object detection.
    
  Thus far, various VoD algorithms have been proposed \cite{seq-nms,tubes,detecttotrack,manet,stsn,psla,selsa,rdn,stmn}.
    In \cite{stmn, stsn, psla},  the visual feature maps obtained for multiple video frames were fused to exploit the temporal redundancy of the representation. 
    FGFA \cite{FGFA} aggregates feature maps using the guidance of the optical-flow map extracted by FlowNet \cite{flownet}.
  STMN \cite{stmn} modifies the ConvGRU module \cite{convgru} to build the spatio-temporal memory and aggregate the spatial features across adjacent frames.
  PSLA \cite{psla} uses the spatial correspondence between adjacent features to align the feature maps for fusion. 
   In \cite{rdn,selsa,mega}, the local features on the objects extracted from each video frame were fused to improve the detection accuracy further.
     STCA \cite{stca} and RDN \cite{rdn} enhance the features in the region proposals by exploring semantic and spatio-temporal relationships among the region proposals. MEGA \cite{mega} employs a spatio-temporal relation module to find the relation between the box-level features over adjacent frames. SELSA \cite{selsa} also aggregates features in region proposals based on the semantic similarity measures.
    However, calculating the similarity measure for all possible pairs of proposals requires high computational complexity.
       In \cite{detecttotrack,manet}, the  features  capturing the motion of objects was extracted from the image sequence and used to find better representation of the objects. 
     D\&T \cite{detecttotrack} predicts bounding box offsets using the correlation map between adjacent frames.
    MANet \cite{manet} extracts motion features based on an optical flow map. 
   
    The proposed TM-VoD method is different from these methods in that it utilizes both temporal redundancy and contextual motion information to improve VoD performance.  The distinct feature of the TM-VoD is that the joint representation of the image sequence and object motion is found by aggregating the aligned and weighted features over two successive stages of object detection and extracting the object motion features effectively both at the pixel level and at the box level.
  

\section{PROPOSED TM-VoD METHOD}
    
    \subsection{Overview}
        The overall architecture of TM-VoD is depicted in Fig. \ref{overall}. TM-VoD consists of three main blocks; 1) {\it temporally gated RPN} (TG-RPN)
        block, 2) \textit{motion-aware temporal box refinement} (MTBR) block, and 3) \textit{joint  temporal  and  motion  feature  generation} (JTMG) block.
        The TG-RPN is performed in the first detection stage and both MTBR and JTMG are performed in the second detection stage.
        
        CNN backbone networks with shared weights are applied to $(M+N+1)$ video frames to generate visual feature maps  $F_{t-M:t+N}=\{F_{t-M},...,F_{t+N}\}$.
        The TG-RPN block produces the enhanced feature map $F_t^{(A)}$ by aggregating $F_{t-M:t+N}$ using the attention weights determined by the {\it gated attention model} (GAM). 
        Based on the feature map $F_t^{(A)}$, the RPN produces the box proposals $b_t^{(A)}$. 
          In addition to temporal feature aggregation, the TG-RPN block extracts  the pixel-level motion feature maps $M_{t-M:t+N}=\{M_{t-M},...,M_{t+N}\}$ based on the incremental changes in the visual feature maps $F_{t-M:t+N}$.  
          
          Next, MTBR block pools the box-level visual and motion features from $F_{t-M:t+N}$ and $M_{t-M:t+N}$. For this goal, the TBOC block predicts the box coordinates $b_{t-M:t+N}$ for $(M+N+1)$  video frames. Instead of adopting computationally-demanding association approach, the MTBR uses $b_t^{(A)}$ as reference boxes (called  {\it anchors}) and predicts the coordinate offsets for  $b_{t-M:t+N}$ based on the features pooled from $F_{t-M:t+N}$ and $M_{t-M:t+N}$.  Then, based on the predicted box coordinates $b_{t-M:t+N}$, the MTBR pools box-level features $r_t^{(A)}$, $r_{t-M:t+N}^{(F)}$, and $r_{t-M:t+N}^{(M)}$ from $F_t^{(A)}$,  $F_{t-M:t+N}$, and  $M_{t-M:t+N}$, respectively. This processing is called {\it RoI alignment}.

         Finally, the JTMB composes the joint representation of the objects using $r_t^{(A)}$,  $r_{t-M:t+N}^{(F)}$, and $r_{t-M:t+N}^{(M)}$. 
        First, $r_{t-M:t+N}^{(F)}$ are temporally aggregated based on the cosine similarity-based attention, which yields box-level aggregated features $g_t^{(F)}$. Second, a {\it bi-directional gated recurrent unit} (bi-GRU) is applied to the sequence of $r_{t-M:t+N}^{(M)}$ to produce box-level motion features $g_t^{(M)}$. Third, the box displacement features $g_t^{(D)}$ are obtained by applying box differential encoding and the fully connected (Fc) layers  to the sequence of the box coordinates $b_{t-M:t+N}$. These three box-level features $g_t^{(M)}$, $g_t^{(D)}$ and $g_t^{(F)}$ are concatenated to perform the final box regression and object classification results. 
    
    \subsection{Temporally Gated RPN (TG-RPN)}
        The first role of TG-RPN is to fuse the visual feature maps $F_{t-M:t+N}$ obtained by CNN backbone networks. To achieve the weighted fusion, the GAM computes the attention weight maps $A_t$ and $A_{t-i}$ and multiplies them to  $F_{t}$ and $F_{t-i}$ as
          \begin{align}
            F_{t-i}^{(G)} = A_{t}\otimes F_{t}+A_{t-i}\otimes F_{t-i},
        \end{align}
          where the operation $\otimes$ denotes pixel-wise multiplication, and  $A_{t}$ and $A_{t-i}$ have channel dimension of 1 and the same spatial size as $F_{t}$ and $F_{t-i}$.  Note that $A_t$ and $A_{t-i}$ gate the contributions of $F_{t}$ and $F_{t-i}$, respectively. $A_t$ and $A_{t-i}$ are obtained from
             \begin{align}
            A_{t} &= \sigma(\mathrm{conv_{3\times3}}(F_{t} \oplus F_{t-i})\\
            A_{t-i} &= 1-A_{t},
        \end{align}
        where $\sigma(\cdot)$ is the logistic-sigmoid function, the operation $\oplus$ denotes concatenation, and ${\mathrm{ conv_{3\times3}}}$ is the convolutional layers with $3 \times 3$ kernels. Note that $A_t$ and $A_{t-i}$ have a value between 0 and 1.  The aggregated feature map $F_{t}^{(A)}$ is obtained from the pixel-wise sum of $F_{t-M}^{G} ... F_{t-1}^{G}, F_{t+1}^{G},  ...,  F_{t+N}^{G}$. Then the box proposals $b_{t}^{(A)}$ for the $t$th frame is obtained from $F_{t}^{(A)}$ using RPN.

        The second role of TG-RPN is to produce the pixel-level motion features $M_{t-M:t+N}$. 
        We employ a motion attention model (MAM), which computes the temporal change of two adjacent visual feature maps, i.e.,  $S_{t-i}=F_{t-i} - F_t$ and applies the channel-wise attention (CWA) of SENet \cite{senet} to $S_{t-i}$
        \begin{align}
               M_{t-i} &= {\rm CWA}(\mathrm{conv_{3\times3}}( S_{t-i})),
        \end{align}
        where global average pooling  is used for CWA.

    \subsection{Motion-aware Temporal Box Reconfiguration (MTBR)}
\begin{figure}[tbh]
        	\centering
            {\includegraphics[width=0.45\textwidth]{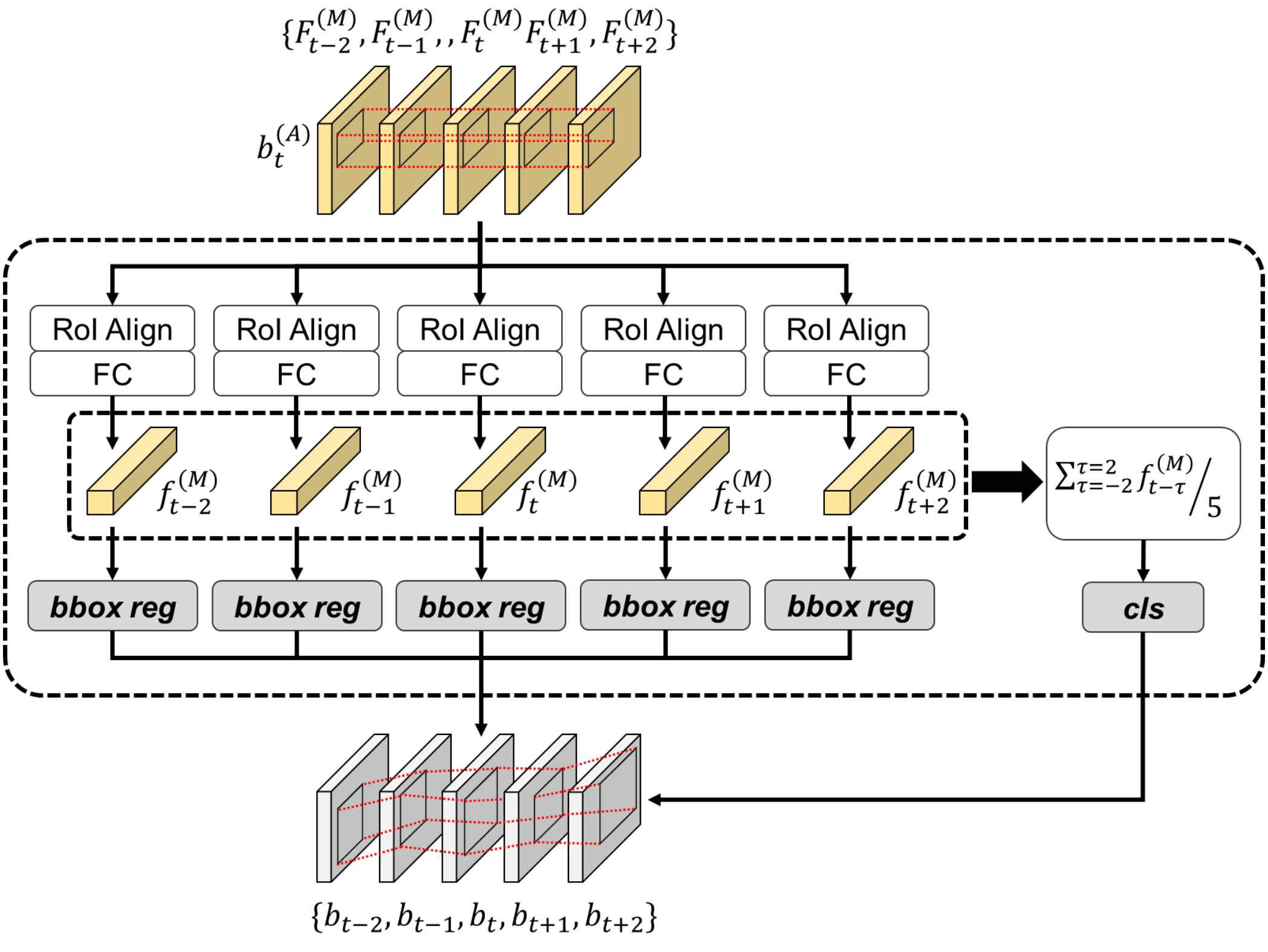}}
        	\caption {\textbf{Structure of TBOC block:} TBOC predicts the relative offsets of the box proposals linked over all video frames. 
      }
        	\label{module}
        \end{figure}

       The role of MTBR is to find the box proposals linked over multiple video frames and pools the box-level features from $F_t^{(A)}$, $F_{t-M:t+N}$, and  $M_{t-M:t+N}$.  Fig. \ref{module}  depicts the structure of the TBOC, which predicts the box proposals $b_{t-M:t+N}$ based on the motion-aware feature maps $F_{t-M:t+N}^{(M)}$ obtained by adding $F_{t-M:t+N}$ and $M_{t-M:t+N}$. As mentioned, the box proposals $b_t^{(A)}$ obtained in the TG-RPN are used as anchors and the box offsets are predicted relative to the anchors for all video frames. Specifically,  the box offset for the $(t+i)$th frame is predicted based on $f_{t+i}^{(M)}$, which is pooled from  $F_{t+i}^{(M)}$ according to $b_t^{(A)}$. Though not being used for the inference step, the object class is also predicted from the element-wise average,  $\sum_{i=-M}^{N}\frac{f_{t+i}^{(M)}}{M+N+1}$  for training purpose only.

       Next, the {\it RoI alignment} block pools the features $r_t^{(A)}$, $r_{t-M:t+N}^{(F)}$, and $r_{t-M:t+N}^{(M)}$ from $F_t^{(A)}$, $F_{t-M:t+N}$, and  $M_{t-M:t+N}$ based on the box proposals $b_{t-M:t+N}$ obtained by the TBOC.
       
         \begin{figure}[tbh]
        	\centering
            {\includegraphics[width=0.45\textwidth]{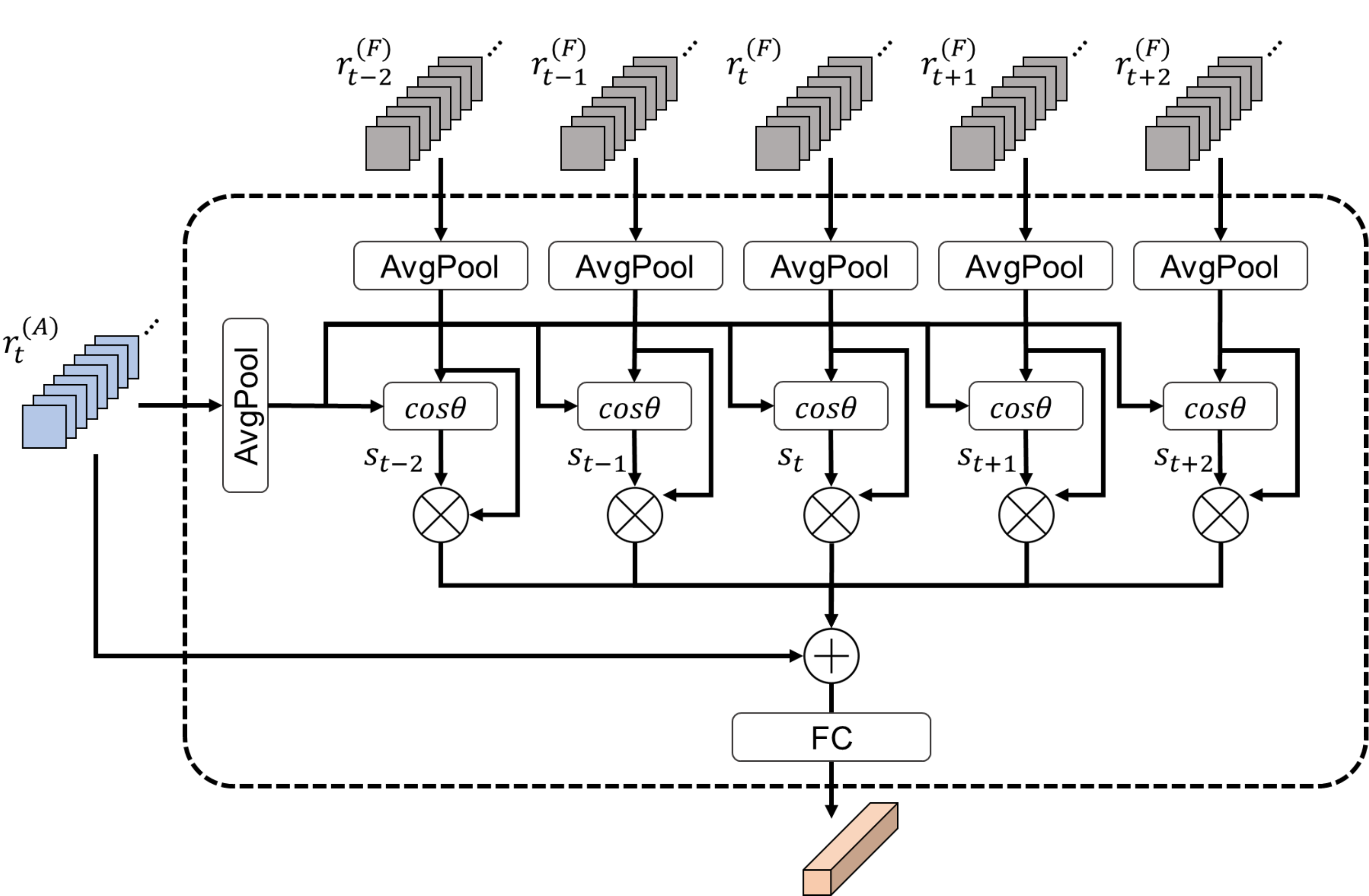}}
        	\caption {\textbf{Structure of box-level gated feature aggregation block:} {\it Box-level gated feature aggregation} weights the RoI-aligned visual features according to the cosine-similarity measure. 
      }
        	\label{module2}
        \end{figure}

    \subsection{Joint Temporal and Motion Feature Generation (JTMG)}
        
        The JTMG block generates three multi-modal box-level features $g_t^{(D)}$, $g_t^{(F)}$, and $g_t^{(M)}$ using the RoI-aligned features $r_t^{(A)}$, $r_{t-M:t+N}^{(F)}$, and $r_{t-M:t+N}^{(M)}$ and combine them to produce the final detection results. 
        
          First, the box-level aggregated features $g_t^{(F)}$ are obtained by fusing the feature maps $r_{t-M:t+N}^{(F)}$ and $r_{t}^{(A)}$ via  \textit{box-level gated feature aggregation} block. As illustrated in Fig.~\ref{module2}, the box-level aggregated feature is expressed as
          \begin{align}
              g_t^{(F)} = \mathrm{fc}(\sum_{i=-M}^{N}w'_{t-i} \phi(r_{t-i}^{(F)}) + \phi(r_{t}^{(A)})),
          \end{align}
          where $\phi(\cdot)$ denotes global average pooling operation and $\mathrm{fc}$ denotes fully connected layer. To obtain the weighted aggregation of $r_{t-M:t+N}^{(F)}$, the attention weight $w'_{t-i}$ is obtained by the cosine similarity between $\phi(r_{t}^{(A)})$ and $\phi(r_{t-i}^{(F)})$, i.e., 
          \begin{align}
                    w'_{t-i}=\frac{ \phi(r_t^{(A)}) \odot \phi(r_{t-i}^{(F)})}{\norm{\phi(r_t^{(A)})}_2 \norm{\phi(r_{t-i}^{(F)})}_2},
              \end{align}
               where $\odot$ denotes the inner product operation.
               Next, the box displacement features $g_t^{(D)}$ are obtained by applying the {\it box differential encoding} to $b_{t-i}$, i.e., $ p_{t-i} = b_{t-i} - b_t$ and passing the encoded box coordinates $p_{t-M:t+N}$ through Fc layers. 
               Third, the box-level motion features $g_t^{(M)}$ are extracted by applying the bi-GRU to $r_{t-M:t+N}^{(M)}$ after converting them into  $1\times 1$ feature vectors using the global average pooling.
         Note that bi-GRU extracts the representation of object motion that captures the temporal change of visual appearance.

        Finally, the box-level object features $g_t^{(D)}$, $g_t^{(M)}$, and $g_t^{(F)}$ are concatenated and used to refine both box regression and object classification results.

    \subsection{Loss Function}
        The multi-task loss function used for the TM-VoD is expressed as
           \begin{align}
        \begin{split}
            L_{\rm total} = \alpha L_{\rm rpn}+ \beta L_{\rm ref}+ \gamma L_{\rm det},
        \end{split}
        \end{align}
        where $L_{\rm rpn}$ consists of the cross-entropy loss for binary object classification task (positive versus negative) and Smooth-L1 loss for box regression task of the RPN \cite{Faster_RCNN}. The parameters $\alpha, \beta$ and $\gamma$ are set to 1.
       The loss term $L_{\rm ref}$ is used to supervise the TBOC. The loss term $L_{\rm ref}$ is expressed as
        \begin{align}
            &L_{\rm ref} = \frac{1}{N_{\rm ref}}\sum_{i=1}^{N_{ref}}{L_{\rm cls|ref}}(i)  \nonumber \\
                 &+   \frac{1}{\sum_{j=t-M}^{t+N} N_{\rm pos|ref}^j}\sum_{j=t-M}^{t+N}\sum_{i=1}^{N_{\rm pos|ref}^j}{L_{\rm reg|ref}(i,j)},
               \end{align}
        where $N_{\rm ref}$ is the number of the box proposals produced by the TBOC at time $t$, $L_{cls|ref}(i)$ denotes the cross-entropy loss for the $i$th box proposal, $N_{\rm pos|ref}^j$ is the number of the positive proposals at time $j$, and $L_{reg|ref}(i,j)$ denotes the Smooth-L1 loss for regression of the $i$th box coordinates at time $j$. 
       The loss term $L_{\rm det}$ comprises the cross-entropy loss for the object classification and smooth-L1 loss for the box regression of the final detection head network \cite{Fast_RCNN}.


\input{./table/ablation_main.tex}
\section{EXPERIMENTS}
    \subsection{Dataset and Evaluation}
       In this section, we present the experimental results to evaluate the performance of the proposed TM-VoD method. We trained the proposed network with both the ImageNet VID dataset and the ImageNet DET dataset following the method suggested in \cite{detecttotrack, FGFA, manet}. The ImageNet DET dataset contains 350k still images with 200 object classes. The ImageNet VID dataset \cite{imagenet} contains 3,862 training videos and 555 validation videos with 30 object classes only. 
       The 30 object classes of the ImageNet VID dataset are a subset of the 200 object classes of the ImageNet DET dataset.
       We subsampled each video by choosing 15 key frames from each video and collected 5 consecutive frames around each key frame. We also collected about 2k images per class from the DET dataset and generated 5 consecutive frames using the same image to generate more class-balanced training data. We tested the VoD methods under consideration on the ImageNet VID validation dataset. We used mean average precision (mAP) metric to evaluate the detection accuracy. We also followed the evaluation protocol in \cite{FGFA}, which divided the objects into three groups, i.e., those with slow, medium, and fast motions. Slow motion refers to the case where intersection over union (IoU) score measured between the present and past frames is higher than 0.9, and fast motion means that the IoU score is lower than 0.7. Medium motion indicates the rest.
        
    \subsection{Implementation Details}
         We used the Faster R-CNN detector \cite{Faster_RCNN} as the baseline network and built our TM-VoD based on it. 
         We used both ResNet-101 \cite{resnet} and ResNeXt-101-32$\times$4d as a backbone network and employed a deformable convolution network \cite{def_conv} to ResNeXt-101.
         Following \cite{detecttotrack}, we first trained the proposed network using the ImageNet DET dataset. Initializing the model with these weights, we finetuned the whole network with
         the ImageNet VID dataset and a part of the ImageNet DET dataset sharing the same 30 object categories with the ImageNet VID dataset. We conducted the data augmentation methods including random flipping, photometric distortion, and  random crop and expansion. 
         The length of the input image sequence was set to 5, i.e., $M=N=2$. The proposed network was trained over 8 epochs with a mini-batch size of 8 on 4 NVIDIA TITAN RTX GPUs. The stochastic gradient descent (SGD) algorithm was used for optimization. The initial learning rate was set to $0.001$ and reduced by a factor of 10 at the 4th epoch and 6th epoch.
         The input images were resized such that the length of a shorter side becomes 600 pixels maintaining the same aspect ratio. 
    \subsection{Ablation Study}
       In this section, we present an ablation study to demonstrate the effectiveness of the ideas used for the proposed method. We compared the performance of the proposed algorithm under the following setups;
        \begin{itemize}
            \item Method (a): Faster R-CNN baseline \cite{Faster_RCNN} with ResNet-101. 
            \item Method (b): The TG-RPN block was added to method (a). The object detection was performed based on the aggregated feature map $F_t^{(A)}$.
            \item Method (c): The MTBR block was added to method (b). Object detection was performed based on the element-wise sum of $r_t^{(A)}, r_{t-2:t+2}^{(F)}$, and $r_{t-2:t+2}^{(M)}$.
            \item Method (d): {\it Box-level gated feature aggregation} was only added to method (c).
            \item Method (e): The JTMG block was added to method (c). 
            \item Method (f): Seq-NMS \cite{seq-nms} was used to post-process the output of the method (e). 
        \end{itemize}
        

         Table \ref{ablation_main} presents an mAP achieved by each method. 
         As a baseline, the method (a) achieves the mAP of $75.45$\%. 
         Applying the gated feature aggregation in TG-RPN, the performance of method (b) improves by  $1.65$\% over that of the method (a). This shows that pixel-level gated feature aggregation yields the enhanced feature maps for object detection. Aligning the object features associated with the box proposals using the the TBOC, the method (c)  achieves a performance gain of $3.72$\% over the baseline detector. This shows that aligning the box-level features for aggregation has a significant impact. 
         The method (d) selectively aggregates the RoI-aligned visual features achieving $0.46$\% improvement over the method (c).
         Using all TG-RPN, MTBR, and JTMG blocks together, the method (e) can achieve up to $5.06$\% performance gain over the baseline, yielding $80.51$\% mAP. The Seq-NMS post-processing offers the further performance improvement from $80.51$\% to $83.07$\%. Note that this amounts to  $7.62$\% improvement over the baseline.

         The mAP performance was also evaluated for slow, medium, fast moving objects.  We observe that   the baseline achieves the worst mAP performance for the fast moving objects. Note that the proposed method offers the largest performance gain ($12.98$\%) for the fast moving objects as compared to $3.24$\% gain for the slow moving objects.

        \begin{figure*}[t]
        	\centering
            \centerline{\includegraphics[width=0.75\textwidth]{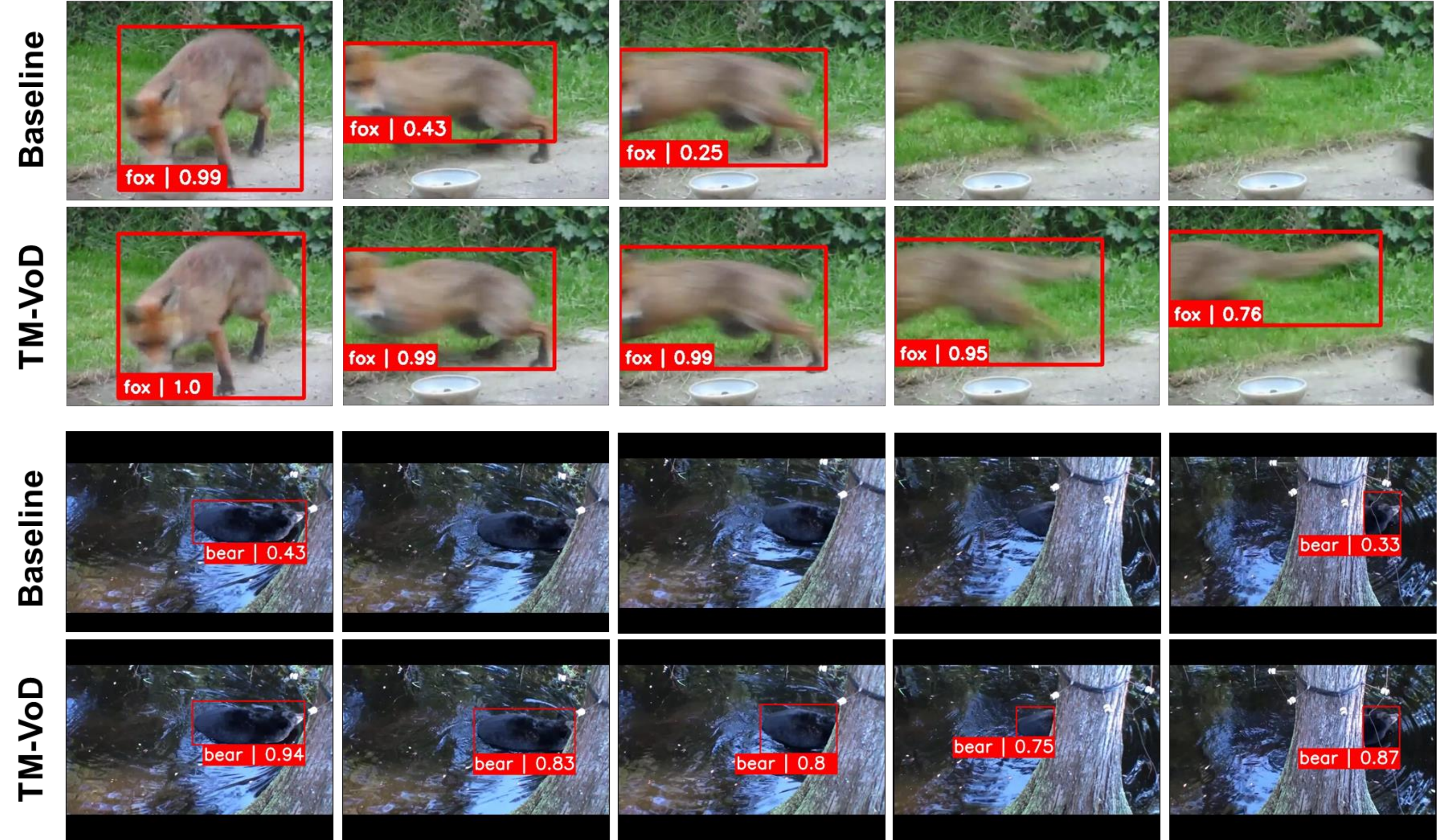}}
        	\caption {\textbf{Comparison of TM-VoD and baseline for the fox and bear videos}: The results of two object detectors (TM-VoD versus Faster R-CNN) are shown for the five video frames of fox and bear videos.  While Faster R-CNN misses the objects for some video frames due to motion blur or occlusion, TM-VoD yields reliable detection results for all five video frames. }
        	\label{box_result}
        \end{figure*}
        \input{./table/sota_seq.tex}

        

    \subsection{Performance Analysis}
    
       Fig.~\ref{box_result} presents the detection results obtained by the baseline (Faster R-CNN) and the TM-VoD for the examples of fox and bear video sequences.  We see that these two video sequences exhibit  motion blur and object occlusion, which prevents the baseline detector from detecting the objects particularly for the degraded images. In contrast, the TM-VoD produces the accurate detection results for all image frames provided.  Even when the detection results from both detectors are correct, the TM-VoD outputs much higher confidence score than the baseline.

        Table \ref{sota_seq} compares the performance of the proposed TM-VoD with the existing VoD methods when evaluated on the ImageNet VID validation set. The performance evaluation is separately performed with and without the post-processing technique.
        Without the post-processing technique, the proposed TM-VoD achieves $83.6$\% mAP, which is higher than other VoD methods  except MEGA \cite{mega}. Note that the MEGA uses 25 consecutive frames as input and thus is required to perform box association for all pairs of box proposals between the adjacent frames.
        When the Seq-NMS is employed as post-processing method, the TM-VoD outperforms all other VoD methods of interest. The detection accuracy achieved by the TM-VoD (i.e., 85.5 \%) is significantly higher than most VoD methods and the performance of the TM-VoD is comparable to that of the current state-of-the-art, MEGA \cite{mega}.



\section{CONCLUSIONS}
In this paper, we presented a novel VoD method, which can produce the joint representation of temporal image sequences and object motion for object detection. First, the hierarchical feature aggregation method was proposed to exploit the temporal redundancy between the video frames. The visual feature maps obtained by the CNN were selectively aggregated using the gated attention weights. In the subsequent detection stage, the box level features were aligned using the box proposals predicted by the TBOC and  were also selectively fused based on the cosine similarity-based weights. Second, the features capturing object motion were obtained in two successive steps. The pixel-level motion feature maps were obtained from the sequence of the visual feature maps. Then, box-level motion features were obtained by applying bi-GRU to the RoI-aligned motion features and extracting the box displacement features. The box-level visual features and motion features were combined to produce the final joint representation of the objects. 
The experiments conducted on ImageNet VID dataset showed that our method achieved the significant performance gain over the baseline and outperformed the existing video object detectors.








\bibliographystyle{plain}
\bibliography{egbib}
\end{document}

%% file: table/ablation_main.tex
\renewcommand{\arraystretch}{1.05}
\begin{table*}[t]
\begin{center}
\begin{adjustbox}{width=0.9\textwidth}

\begin{tabular}{c|c|c|c|c|c|c|c}
\Xhline{4\arrayrulewidth}
\multicolumn{2}{c|}{} & \multicolumn{6}{c}{Faster R-CNN with ResNet-101} \\ \hline
\multicolumn{2}{c|}{Methods} & Method (a) & Method (b) & Method (c) & Method (d) & Method (e) & Method (f) \\ \hline\hline

\multicolumn{2}{c|}{TG-RPN} & & \checkmark & \checkmark & \checkmark & \checkmark &\checkmark \\ \cline{1-2}

\multicolumn{2}{c|}{MTBR} & &            & \checkmark & \checkmark  & \checkmark & \checkmark \\ \cline{1-2}

\multirow{3}{*}{JTMG} & Box-level gated feature aggregation & & & & \checkmark  & \checkmark  & \checkmark \\

     & Box displacement feature extractor & & & & & \checkmark & \checkmark \\
     
      & Box-level motion feature extractor & & & & & \checkmark & \checkmark \\ \cline{1-2}
      
\multicolumn{2}{c|}{Post-processing}                           & & & & & & \checkmark \\ \hline

\multicolumn{2}{c|}{mAP (\%)}          & 75.45 & 77.10$_{\uparrow 1.65}$   & 79.17$_{\uparrow 3.72}$ & 79.63$_{\uparrow 4.18}$& 80.51$_{\uparrow 5.06}$ & 83.07$_{\uparrow 7.62}$  \\ \hline
\multicolumn{2}{c|}{mAP (\%) (slow)}   & 84.37 & 84.64$_{\uparrow 0.27}$ & 84.79$_{\uparrow 0.42}$ & 85.20$_{\uparrow 0.83}$& 85.54$_{\uparrow 1.17}$ & 87.86$_{\uparrow 3.24}$   \\ \hline
\multicolumn{2}{c|}{mAP (\%) (medium)} & 73.38 & 75.44$_{\uparrow 2.06}$   & 78.04$_{\uparrow 4.63}$ & 78.65$_{\uparrow 5.27}$& 79.96$_{\uparrow 6.58}$ & 81.87$_{\uparrow 8.49}$    \\ \hline
\multicolumn{2}{c|}{mAP (\%) (fast)}   & 53.67 & 56.59$_{\uparrow 2.92}$   & 60.98$_{\uparrow 7.31}$ & 61.81$_{\uparrow 8.14}$& 63.65$_{\uparrow 9.98}$ & 66.65$_{\uparrow 12.98}$   \\ \hline

\Xhline{4\arrayrulewidth}

\end{tabular}

\end{adjustbox}
\end{center}
\caption{Ablation study conducted on the ImageNet VID validation set}

\label{ablation_main}
\end{table*}
\renewcommand{\arraystretch}{1.05}

%% file: table/sota_seq.tex
\renewcommand{\arraystretch}{1.0}
\begin{table}[t]
\begin{center}
\begin{adjustbox}{width=0.90\linewidth}
\begin{Large}
\begin{tabular}{c | c | c | c }
\Xhline{4\arrayrulewidth}

Network & Backbone & Post-Processing  & mAP (\%) \\ \hline\hline

D\&T\cite{detecttotrack} & Inception v-4                  & - & 82.0\\
FGFA \cite{FGFA}         & ResNet-101                     & - & 76.3\\
MANet \cite{manet}        & ResNet-101                     & - & 78.1\\
RDN \cite{rdn}           & ResNeXt-101        & - & 83.2\\
SELSA \cite{selsa}       & ResNeXt-101      & - & 83.1\\
MEGA \cite{mega}         & ResNeXt-101     & - & \textbf{84.5}\\
\textbf{Ours}            & ResNeXt-101       & - & 83.6\\ \hline\hline

D\&T \cite{detecttotrack} & Inception v-4                  & Viterbi & 82.1\\
FGFA \cite{FGFA}         & ResNet-101                     & Seq-NMS & 78.4\\
MANet\cite{manet}        & ResNet-101                     & Seq-NMS & 80.3\\
STMN \cite{stmn}         & ResNet-101                     & Seq-NMS & 80.5\\
RDN \cite{rdn}           & ResNeXt-101       & BLR     & 84.7\\
SELSA \cite{selsa}       & ResNeXt-101       & Seq-NMS & 83.7\\
MEGA \cite{mega}         & ResNeXt-101       & BLR & 85.4\\
\textbf{Ours}            & ResNeXt-101       & Seq-NMS & \textbf{85.5}\\

\Xhline{4\arrayrulewidth}

\end{tabular}
\end{Large}
\end{adjustbox}
\end{center}
\caption{Performance comparison of several VoD methods
evaluated on the ImageNet VID validation set
}

\label{sota_seq}
\end{table}
\renewcommand{\arraystretch}{1}

%% file: root.bbl
\begin{thebibliography}{10}

\bibitem{convgru}
Nicolas Ballas, Li~Yao, Chris Pal, and Aaron Courville.
\newblock Delving deeper into convolutional networks for learning video
  representations.
\newblock {\em arXiv preprint arXiv:1511.06432}, 2015.

\bibitem{stsn}
Gedas Bertasius, Lorenzo Torresani, and Jianbo Shi.
\newblock Object detection in video with spatiotemporal sampling networks.
\newblock In {\em Proceedings of the European Conference on Computer Vision
  (ECCV)}, pages 331--346, 2018.

\bibitem{mega}
Yihong Chen, Yue Cao, Han Hu, and Liwei Wang.
\newblock Memory enhanced global-local aggregation for video object detection.
\newblock In {\em Proceedings of the IEEE/CVF Conference on Computer Vision and
  Pattern Recognition}, pages 10337--10346, 2020.

\bibitem{RFCN}
Jifeng Dai, Yi~Li, Kaiming He, and Jian Sun.
\newblock R-fcn: Object detection via region-based fully convolutional
  networks.
\newblock In {\em Advances in neural information processing systems}, pages
  379--387, 2016.

\bibitem{def_conv}
Jifeng Dai, Haozhi Qi, Yuwen Xiong, Yi~Li, Guodong Zhang, Han Hu, and Yichen
  Wei.
\newblock Deformable convolutional networks.
\newblock In {\em Proceedings of the IEEE international conference on computer
  vision}, pages 764--773, 2017.

\bibitem{rdn}
Jiajun Deng, Yingwei Pan, Ting Yao, Wengang Zhou, Houqiang Li, and Tao Mei.
\newblock Relation distillation networks for video object detection.
\newblock {\em arXiv preprint arXiv:1908.09511}, 2019.

\bibitem{flownet}
Alexey Dosovitskiy, Philipp Fischer, Eddy Ilg, Philip Hausser, Caner Hazirbas,
  Vladimir Golkov, Patrick Van Der~Smagt, Daniel Cremers, and Thomas Brox.
\newblock Flownet: Learning optical flow with convolutional networks.
\newblock In {\em Proceedings of the IEEE international conference on computer
  vision}, pages 2758--2766, 2015.

\bibitem{detecttotrack}
Christoph Feichtenhofer, Axel Pinz, and Andrew Zisserman.
\newblock Detect to track and track to detect.
\newblock In {\em Proceedings of the IEEE International Conference on Computer
  Vision}, pages 3038--3046, 2017.

\bibitem{Fast_RCNN}
Ross Girshick.
\newblock Fast r-cnn.
\newblock In {\em Proceedings of the IEEE international conference on computer
  vision}, pages 1440--1448, 2015.

\bibitem{RCNN}
Ross Girshick, Jeff Donahue, Trevor Darrell, and Jitendra Malik.
\newblock Rich feature hierarchies for accurate object detection and semantic
  segmentation.
\newblock In {\em Proceedings of the IEEE conference on computer vision and
  pattern recognition}, pages 580--587, 2014.

\bibitem{tubes}
Georgia Gkioxari and Jitendra Malik.
\newblock Finding action tubes.
\newblock In {\em Proceedings of the IEEE conference on computer vision and
  pattern recognition}, pages 759--768, 2015.

\bibitem{psla}
Chaoxu Guo, Bin Fan, Jie Gu, Qian Zhang, Shiming Xiang, Veronique Prinet, and
  Chunhong Pan.
\newblock Progressive sparse local attention for video object detection.
\newblock {\em arXiv preprint arXiv:1903.09126}, 2019.

\bibitem{seq-nms}
Wei Han, Pooya Khorrami, Tom~Le Paine, Prajit Ramachandran, Mohammad
  Babaeizadeh, Honghui Shi, Jianan Li, Shuicheng Yan, and Thomas~S Huang.
\newblock Seq-nms for video object detection.
\newblock {\em arXiv preprint arXiv:1602.08465}, 2016.

\bibitem{Mask_RCNN}
Kaiming He, Georgia Gkioxari, Piotr Doll{\'a}r, and Ross Girshick.
\newblock Mask r-cnn.
\newblock In {\em Proceedings of the IEEE international conference on computer
  vision}, pages 2961--2969, 2017.

\bibitem{resnet}
Kaiming He, Xiangyu Zhang, Shaoqing Ren, and Jian Sun.
\newblock Deep residual learning for image recognition.
\newblock In {\em Proceedings of the IEEE conference on computer vision and
  pattern recognition}, pages 770--778, 2016.

\bibitem{mobilenet}
Andrew~G Howard, Menglong Zhu, Bo~Chen, Dmitry Kalenichenko, Weijun Wang,
  Tobias Weyand, Marco Andreetto, and Hartwig Adam.
\newblock Mobilenets: Efficient convolutional neural networks for mobile vision
  applications.
\newblock {\em arXiv preprint arXiv:1704.04861}, 2017.

\bibitem{senet}
Jie Hu, Li~Shen, and Gang Sun.
\newblock Squeeze-and-excitation networks.
\newblock In {\em Proceedings of the IEEE conference on computer vision and
  pattern recognition}, pages 7132--7141, 2018.

\bibitem{alexnet}
Alex Krizhevsky, Ilya Sutskever, and Geoffrey~E Hinton.
\newblock Imagenet classification with deep convolutional neural networks.
\newblock In {\em Advances in neural information processing systems}, pages
  1097--1105, 2012.

\bibitem{RetinaNet}
Tsung-Yi Lin, Priya Goyal, Ross Girshick, Kaiming He, and Piotr Doll{\'a}r.
\newblock Focal loss for dense object detection.
\newblock In {\em Proceedings of the IEEE international conference on computer
  vision}, pages 2980--2988, 2017.

\bibitem{SSD}
Wei Liu, Dragomir Anguelov, Dumitru Erhan, Christian Szegedy, Scott Reed,
  Cheng-Yang Fu, and Alexander~C Berg.
\newblock Ssd: Single shot multibox detector.
\newblock In {\em European conference on computer vision}, pages 21--37.
  Springer, 2016.

\bibitem{stca}
Hao Luo, Lichao Huang, Han Shen, Yuan Li, Chang Huang, and Xinggang Wang.
\newblock Object detection in video with spatial-temporal context aggregation.
\newblock {\em arXiv preprint arXiv:1907.04988}, 2019.

\bibitem{YOLO}
Joseph Redmon, Santosh Divvala, Ross Girshick, and Ali Farhadi.
\newblock You only look once: Unified, real-time object detection.
\newblock In {\em Proceedings of the IEEE conference on computer vision and
  pattern recognition}, pages 779--788, 2016.

\bibitem{Faster_RCNN}
Shaoqing Ren, Kaiming He, Ross Girshick, and Jian Sun.
\newblock Faster r-cnn: Towards real-time object detection with region proposal
  networks.
\newblock In {\em Advances in neural information processing systems}, pages
  91--99, 2015.

\bibitem{imagenet}
Olga Russakovsky, Jia Deng, Hao Su, Jonathan Krause, Sanjeev Satheesh, Sean Ma,
  Zhiheng Huang, Andrej Karpathy, Aditya Khosla, Michael Bernstein, et~al.
\newblock Imagenet large scale visual recognition challenge.
\newblock {\em International journal of computer vision}, 115(3):211--252,
  2015.

\bibitem{vgg}
Karen Simonyan and Andrew Zisserman.
\newblock Very deep convolutional networks for large-scale image recognition.
\newblock {\em arXiv preprint arXiv:1409.1556}, 2014.

\bibitem{manet}
Shiyao Wang, Yucong Zhou, Junjie Yan, and Zhidong Deng.
\newblock Fully motion-aware network for video object detection.
\newblock In {\em Proceedings of the European Conference on Computer Vision
  (ECCV)}, pages 542--557, 2018.

\bibitem{selsa}
Haiping Wu, Yuntao Chen, Naiyan Wang, and Zhaoxiang Zhang.
\newblock Sequence level semantics aggregation for video object detection.
\newblock {\em arXiv preprint arXiv:1907.06390}, 2019.

\bibitem{stmn}
Fanyi Xiao and Yong Jae~Lee.
\newblock Video object detection with an aligned spatial-temporal memory.
\newblock In {\em Proceedings of the European Conference on Computer Vision
  (ECCV)}, pages 485--501, 2018.

\bibitem{FGFA}
Xizhou Zhu, Yujie Wang, Jifeng Dai, Lu~Yuan, and Yichen Wei.
\newblock Flow-guided feature aggregation for video object detection.
\newblock In {\em Proceedings of the IEEE International Conference on Computer
  Vision}, pages 408--417, 2017.

\end{thebibliography}
